\documentclass[letterpaper, 10 pt, conference]{ieeeconf}  

\IEEEoverridecommandlockouts                              

\usepackage{cite}
\usepackage{amsmath,amssymb,amsfonts}
\usepackage{algorithmic}
\usepackage{graphicx}
\usepackage{textcomp}
\usepackage{xcolor}
\usepackage[subrefformat=parens]{subcaption}
\usepackage{multirow}
\usepackage{hyperref}
\usepackage{siunitx}

\title{\LARGE \bf
AutoMCM: Maneuver Coordination Service with Abstracted Functions for Autonomous Driving}

\author{Masaya Mizutani$^{1}$, Manabu Tsukada$^{1}$, and Hiroshi Esaki$^{1}$
\thanks{$^{1}$ Graduate School of Information Science and Technology, The University of Tokyo, Tokyo, Japan
        {\tt\small mizu@hongo.wide.ad.jp, tsukada@hongo.wide.ad.jp, hiroshi@wide.ad.jp}}%
}

\begin{document}

\maketitle
\thispagestyle{empty}
\pagestyle{empty}

\begin{abstract}
        A cooperative intelligent transport system (C-ITS) uses vehicle-to-everything (V2X) technology to make self-driving vehicles safer and more efficient. Current C-ITS applications have mainly focused on real-time information sharing, such as for cooperative perception. 
        In addition to better real-time perception, self-driving vehicles need to achieve higher safety and efficiency by coordinating action plans. 
        This study designs a maneuver coordination (MC) protocol that uses seven messages to cover various scenarios and an abstracted MC support service. We implement our proposal as AutoMCM by extending two open-source software tools: Autoware for autonomous driving and OpenC2X for C-ITS. The results show that our system effectively reduces the communication bandwidth by limiting message exchange in an event-driven manner. Furthermore, it shows that the vehicles run 15\% faster when the vehicle speed is 30 km/h and 28\% faster when the vehicle speed is 50 km/h using our scheme. Our system shows robustness against packet loss in experiments when the message timeout parameters are appropriately set.
\end{abstract}

\section{Introduction}
In recent years, there has been a great deal of research and development related to automated driving.
A key focus area has been the enhancement of traffic safety and improved passenger comfort. 
Autonomous driving is performed using sensors for object detection, path planning, and actuation. Research to improve safety and efficiency is being developed for each process.
Various organizations are carrying out autonomous automatic driving development, and several open-source software packages have been developed, including Autoware~\cite{kato2018autoware} and Apollo~\cite{apollo:github}. 
%

A cooperative intelligent transport system (C-ITS)~\cite{cits} uses vehicle-to-everything (V2X) technology to make driving safer and more efficient. Hence, it has attracted the attention of researchers' and engineers' worldwide. 
European Telecommunications Standards Institute (ETSI) and International Organization for Standardization (ISO) have issued standards to promote technological deployment in this area. 
C-ITS defines network architectures and messaging for V2X.
For example, the cooperative awareness message (CAM)~\cite{ETSI-EN-302-637-2-CAM} is used by vehicles to provide real-time information about its status. The collective perception message (CPM)~\cite{cpm} is used to share real-time information about the vehicle's surroundings.
Unfortunately, these messages are not designed for cooperative path planning, which requires the inclusion of future (planned) information for overall optimal path planning.

The future information comprises a list of the vehicle's planned positions over time (i.e., the planned trajectory). 
ETSI defines a maneuver coordination message (MCM) as one that exchanges planned trajectories and performs driving coordination.
However, the MCM format is still under development~\cite{ETSI-TR-103-578-MCM}, and no standard has been issued.

The contributions of this study are as follows.
We organize MCM requirements and propose one that meets all extant requirements.
We design a maneuver coordination service for various scenarios and autonomous driving applications, including adaptive cruise control (ACC) and fully autonomous driving. We implement the proposed maneuver coordination system (i.e., AutoMCM) by extending the open-source software, Autoware and OpenC2X~\cite{OpenC2X:website}. To the best of our knowledge, this is the first work to realize a maneuver coordination service by integrating Autoware and OpenC2X.

The rest of the paper is organized as follows.
Section~\ref{sec:related} presents research related to maneuver coordination.
Section~\ref{sec:issues} summarizes the issues, the requirements, and our approaches. 
In Section~\ref{sec:proposed}, we design a maneuver coordination service with abstracted functions compliant with the intelligent transport system (ITS) station architecture per ISO and ETSI. 
Section~\ref{sec:implementation} describes our AutoMCM implementation in detail.
Section~\ref{sec:evaluation} shows the results of simulator experiments. Finally, Section~\ref{sec:conclusions} presents conclusions and potential future works.

\section{Related works}\label{sec:related}
Research on maneuver coordination can be categorized into two categories: scenario-specific MC and general MC.
Scenario-specific maneuver coordination supports platooning, lane changing, and merge coordination; several field experiments are in progress.
However, research on general-purpose maneuver coordination is still in its infancy. For example, few researchers have yet evaluated the impact of network delays and packet losses on driving comfort.

In a study on platooning coordination, a V2X protocol to deal with all possible platooning scenarios is proposed~\cite{englund2016grand}.
The protocol coordinates vehicles to merge two platooning sequences into one.
Additionally, in \cite{Sawade2018-rj} the authors proposed a redundant message protocol for cooperative driving involving several vehicles, such as platooning.
In \cite{Milanes2011-bl}, the roadside unit at the merging point detects the merging activity, searches for vehicles on the main road that may be involved in a collision, and subsequently instructs the identified vehicles to adjust their speeds.
In \cite{Eiermann2020-np}, the cooperative maneuver protocol (CMP) as a coordination message at the merge point is proposed. 
Moreover, the lane-change messaging for coordinated lane changing is proposed in \cite{Wang2018-ef}. 
Several studies have used the space-time reservation protocol (STRP), which uses messaging to reserve a specific location for a certain amount of time. In the work of \cite{Heb2018-dx}, STRP was used for lane-change coordination, and field experiments were conducted to validate the technique. 
The extended method to intersections, passing, and roundabouts is proposed in \cite{Nichting2020-zy}. 

The TransAID project~\cite{Schindler2019-lc} performs research and development on more generic V2X messages. 
In this project, cooperative automated driving and C-ITS standard messages such as CAM, CPM, and MCM have been studied~\cite{correa2019infrastructure, correa2019transaid}. 
A more generic MCM message format is proposed in \cite{correa2019infrastructure}, which considered transition of control (switching from automatic to manual driving) as a type of instruction for MCM in addition to lane change and speed adjustment.
A V2X framework to exchange each message sets standardized in C-ITS is implemented in \cite{correa2019transaid}.

A full-stack from the access layer to the facility layer is implemented in \cite{Jacob_undated-ga, auerswald2019cooperative}. 
Alongside MCM, a maneuver recommended message was used for driving coordination.
In \cite{Hafner2020-xn},  the complex vehicular interactions protocol for driving coordination is proposed. 
In \cite{Decentra11:online} it is assumed that all vehicles always deliver a trajectory when performing driving cooperation.
In \cite{Lehmann2018-bb},  MCM is used to reduce uncertainty in automated vehicles and proposed the MCM as a more general protocol. 
In our previous works~\cite{Tsukada2020-ib, Tsukada2020b}, we have implemented AutoC2X, which enable cooperative perception on Autoware and OpenC2X.

\section{Issues and approaches}\label{sec:issues}
In this research, maneuver coordination is achieved by exchanging trajectories to support more general scenarios and applications.
Based on this premise, the following subsections describe design policies to solve maneuver coordination issues. 

\subsection{Bandwidth saving and robustness}
Because messages containing trajectories require a large number of data, their distribution increases network load.
Therefore, we designed an event-driven maneuver coordination service to conserve bandwidth. 
When coordination is needed, an initiator triggers a message to begin coordination. 
This avoids unnecessary data exchange.
Furthermore, because the MCM only needs to convey a message once, the sender continues to send it until the acknowledgment is received.
This scheme allows the message to be sent correctly even in unreliable communication environments while conserving bandwidth.

\subsection{Multiple scenario adaptability}
Multiple scenarios require maneuver coordination, such as lane changing and intersection navigation, and the applications of automated driving vary from ACC to fully automated driving.
Therefore, it is desirable to design a system architecture to support a variety of scenarios and applications.
We divided the maneuver coordination functions into two parts: common functions and scenario-specific functions. We placed the common functions on the facility layer and scenario-specific functions in the application layer in the ITS station architecture~\cite{ISO-21217-CALM-Arch, ETSI-EN-302-665-Arch} in ETSI and ISO. 
This design allows a system implementation that can handle generic scenarios and applications.

\subsection{Safety}
As mentioned in \cite{Lehmann2018-bb}, instructions can be duplicated. 
A vehicle providing instructions will be unaware that the receiving vehicle also has received another vehicle's instructions in real scenarios. In this case, original instructions may be overwritten, creating a potentially dangerous situation. 
Therefore, we designed a state management scheme concerning maneuver coordination that ensures that if a vehicle is in the middle of maneuver coordination and a message from another vehicle comes in, the receiving vehicle can refuse the instruction to avoid an unexpected overwrite of instructions.

\subsection{Traffic comfort}
During maneuver coordination, the traffic scenario may change according to the surrounding environment, or the vehicle may malfunction.
Without mitigation, a vehicle that cannot handle the scenario will continue to decelerate unnecessarily, which can obviously become problematic.
Thus, we designed a scheme to detect abnormal situations during maneuver coordination and return vehicles to stand-alone autonomous driving if necessary. 
This scheme allows, at a minimum, safe driving during abnormal situations and may improve traffic comfort by avoiding unnecessary deceleration when a scenario changes.

\section{Proposed maneuver coordination service}\label{sec:proposed}

\subsection{System architecture}
\begin{figure}[tbp]
  \centering
  \includegraphics[width=\linewidth]{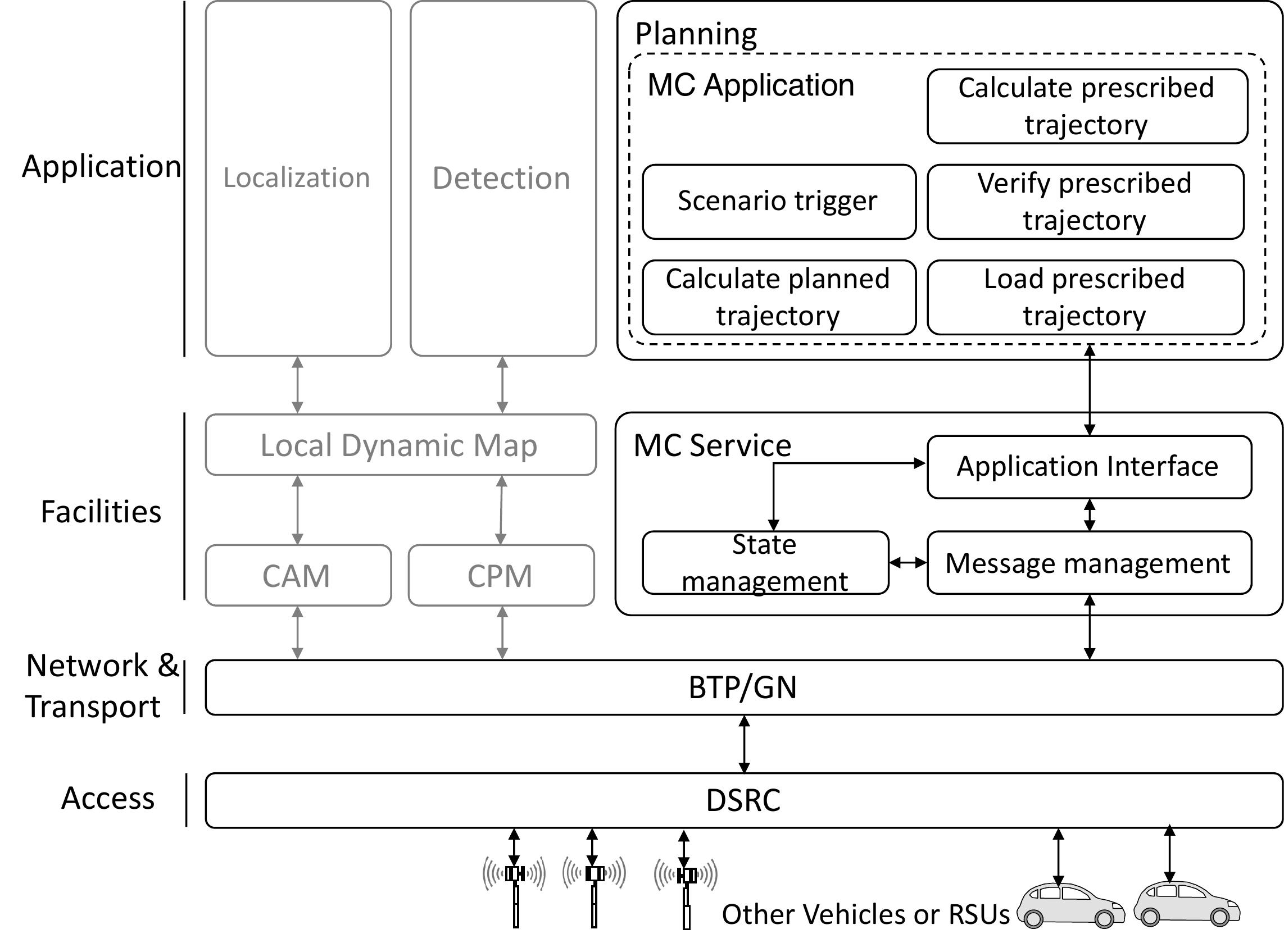}
  \caption{System architecture}
  \label{fig:system_arch}
\end{figure}

Figure~\ref{fig:system_arch} presents an overview of the system architecture.
In this study, we comply with the C-ITS standard architecture of the ISO and ETSI. 
As mentioned in the requirements, this research proposes a system architecture used in generic scenarios and applications. We designed the abstracted functions in the facility layer and the scenario-specific functions in the application layer. 
The \emph{MC service} is commonly required for all scenarios and applications integrated into the facilities. 
This service provides functions of the application interface, message generation, transmission and reception, and state management.
On the other hand, the \emph{MC application} is specific to each scenario located in the application layer.
The MC application calculates each trajectory, triggers scenarios, and verifies and loads prescribed trajectories.

\subsection{MCM Flow}
\begin{figure}[tbp]
  \centering
  \includegraphics[width=\linewidth]{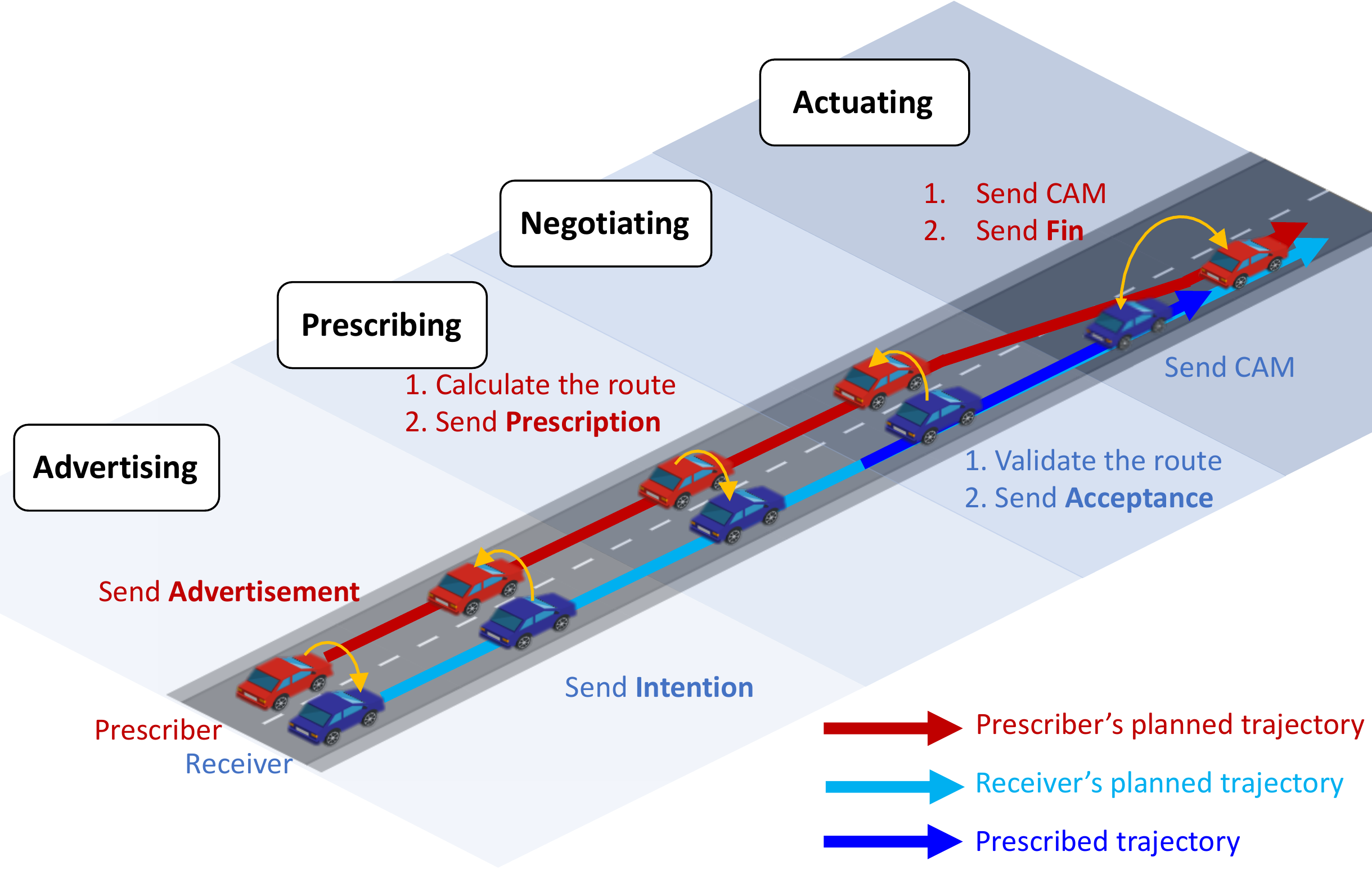}
  \caption{Proposed MCM flow}
  \label{fig:mcm_flow}
\end{figure}

Figure~\ref{fig:mcm_flow} shows the proposed MCM flow in a lane-change scenario. 
The prescriber vehicle changes lanes, and a receiver vehicle is in the target lane.
The \textbf{planned trajectory} is the vehicles' future trajectory maintained in the autonomous driving system. The
\textbf{prescribed trajectory} is the trajectory transmitted from a prescriber to instruct the receiver's maneuver. 
We define the following messages to describe our proposition:
\begin{itemize}
  \item \textbf{Advertisement} is the first message sent by a vehicle that wants to perform maneuver coordination on another in various scenarios, such as lane changing or merge coordination. The message contains the scenario.
  \item \textbf{Intention} is a reply message to \emph{Advertisement} with the planned trajectory. It also contains the target station identifier (ID) to clarify to which vehicle the message is sent.
  \item \textbf{Prescription} is a message that gives instructions to the vehicle that returned the Intention. It contains a \emph{Prescribed Trajectory} to provide instructions. It contains the target station ID to specify the destination. 
  \item \textbf{Acceptance} is a message that indicates whether or not to allow the \emph{Prescription}.  It contains the target station ID, acceptance (whether or not to allow), and selected trajectory.
  \item \textbf{Fin} is a message to indicate the end of maneuver coordination.
  \item \textbf{Cancel} is a message to quit the scenario in the middle of maneuver coordination.
  \item \textbf{Ack} is a message to confirm that the message has been received.
\end{itemize}

The following subsections describe the flow of messages and state transitions.

\subsubsection{Advertising}
First, when the prescriber desires a lane change, it transitions to an advertising state and distributes an \emph{Advertisement} to the surrounding area for a certain period.
Then, any receivers reply to the \emph{Intention} that contains the receiver's planned trajectory to the prescriber.
When the prescriber receives the \emph{Intention}, it replies with an \emph{Ack} to inform the receiver that it has received the \emph{Intention}.
The receiver continues to send the \emph{Intention} until the \emph{Ack} is returned.

\subsubsection{Prescribing}
The prescriber transmits its state of prescription after a certain period, and judges whether there is a risk of collision between the prescriber's planned trajectory and the receivers' planned trajectories based on the received \emph{Intention}.
If there is a risk, the prescriber calculates a prescribed trajectory that avoids the collision and sends the new \emph{ Prescription} containing it to the receiver.
The prescription also confirms the response as an \emph{Ack}.

\subsubsection{Negotiating}
When the receiver receives the \emph{Prescription}, it checks whether it is safe to drive according to the prescribed trajectory.
If it is safe, the receiver sends the \emph{Acceptance} to the prescriber.
If it is not accepted, the prescriber returns to the state of prescribing again and recalculates the prescribed trajectory.
The \emph{Acceptance} also confirms the response as an \emph{Ack}.

\subsubsection{Actuating}
Finally, the receiver executes the prescribed trajectory.
During this time, both vehicles transmit CAMs at regular intervals.
If a CAM cannot be received for a certain period, both vehicles consider it a communication error, stop this scenario, and return to stand-alone autonomous driving.
After successfully executing the coordination to the end, the prescriber sends the \emph{Fin} and terminates the scenario.

\section{AutoMCM implementation} \label{sec:implementation}
\subsection{System design}
\subsubsection{System model}
The proposed method was implemented in this study by extending and integrating two open-source software packages (i.e., Autoware and OpenC2X). Fig~\ref{fig:implementation} shows the proposed system model with the implementation details. 
OpenC2X is in charge of MC services in the facility layer, and Autoware supports the functions in the application layer. 
We enabled the two software packages to communicate with each other using the JavaScript Object Notation format\cite{rfc8259} using Websocket~\cite{rfc6455}. 
This design allows users to have a flexible system configuration by installing the software on different computers, such as a host and a router. 
The recommended configuration in \cite{ISO-21217-CALM-Arch} holds that a router is in charge of external connectivity to a group of hosts connected to the in-vehicle wired network. 
In our design, the hosts can access the facility layer function via the in-vehicle network.

\subsubsection{Autoware}

Autoware~\cite{kato2018autoware} was implemented on a middleware robot operating system (ROS)~\cite{Quigley2009-eu} for robot control and provided the necessary functions for automated driving. 
To perform distributed processing, each process was implemented as a \emph{node}. Each node exchange information on \emph{topics}. 
ROS also includes the 3-dimensional (3D) \emph{RViz} visualization tool, which shows the status of tasks. 
ROS provides a Websocket server (i.e., ROS Bridge) as a standard feature, and we used this server to connect with OpenC2X. 

Autoware provides a set of applications necessary for autonomous driving, such as localization, perception, planning, and control from the input of 3D maps using Lanelet2~\cite{Poggenhans2018-fk}, LiDAR, a camera, and the Global Navigation Satellite System. The perception module provides the detected obstacle information to the planning module.  
Autoware has multiple planning modules for various scenarios because it is technically challenging to have a unified planner for all situations. 
The \emph{lane-driving planner} has the functions necessary for a regular road, including normal driving and lane change. 
The \emph{behavior planner} determines lane changes and turn-signal activations. The \emph{motion planner} is responsible for optimizing the shape of the trajectory with a given lateral acceleration and jerk limit. 
Autoware also provides a \emph{planning simulator} that performs experiments of the planning module by simulation.
The planning module gives the trajectory to the \emph{control} module, which manipulates the acceleration, braking, and steering of the autonomous vehicle via a controller area-network controller.

\subsubsection{OpenC2X}
In OpenC2X, the ITS station architecture is implemented except for the security layer. 
The ITS GeoNetworking (GN) module~\cite{ETSI-EN-302-636-4-1-GN} is only implemented to add the GN header to packets, but the GN routing has not yet been implemented.
In OpenC2X, CAMs and decentralized environmental notification messages are generated based on the information obtained from onboard diagnostics and the Global Positioning System, and the information is stored in a local dynamic map ~\cite{ETSI-EN-302-895-LDM} via a single-hop broadcast with decentralized congestion control.
On the other hand, there is no MCM implementation, which has not yet been standardized.

\subsection{Implementation overview}
\begin{figure}[htb]
  \centering
  \includegraphics[width=\linewidth]{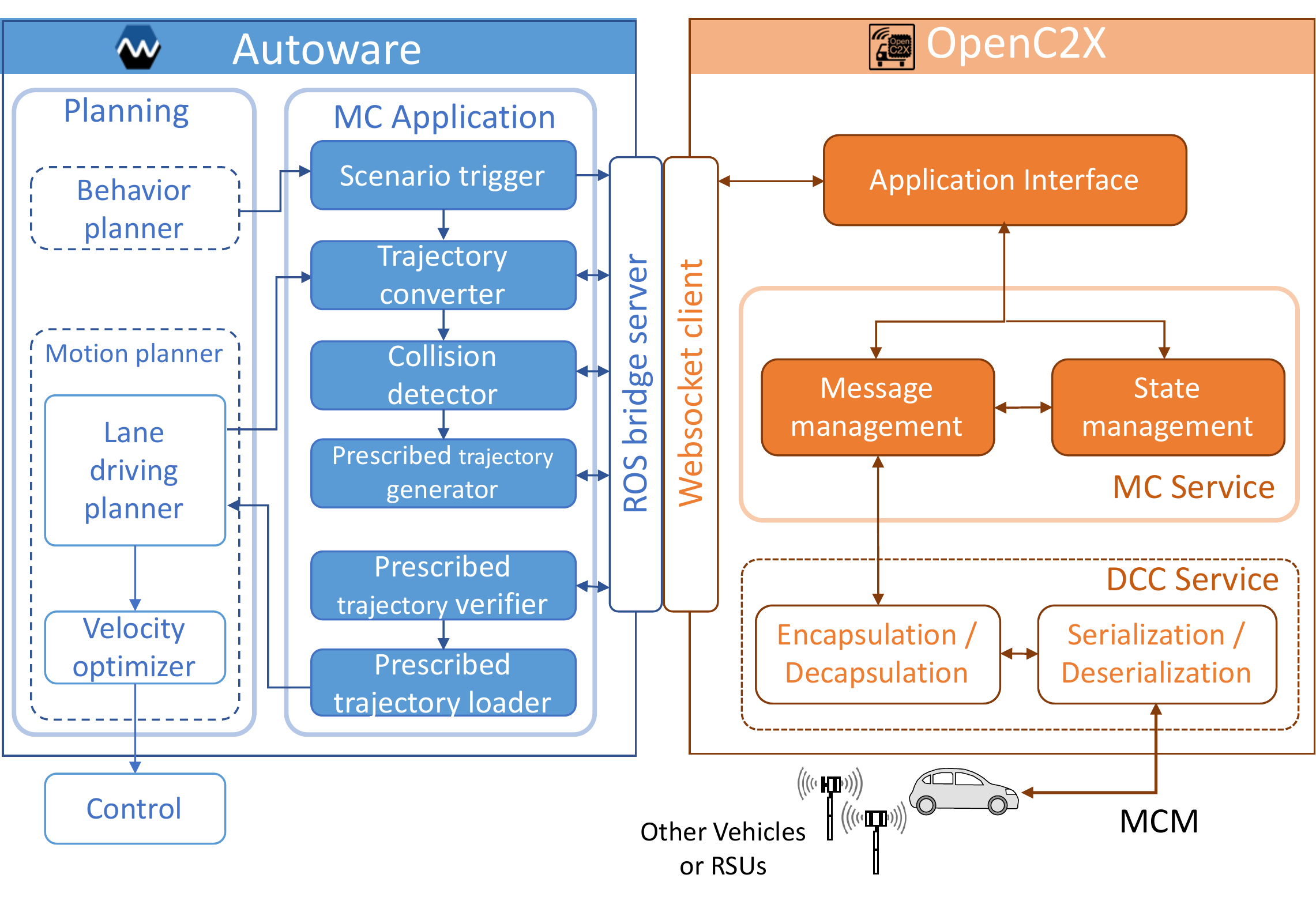}
  \caption{Implementation architecture}
  \label{fig:implementation}
\end{figure}

An overview of the implementation architecture is presented in Fig.~\ref{fig:implementation}. We used Autoware.IV v0.8.0 and OpenC2X-standalone v1.5 for the implementation. 
The white boxes represent the existing implementations, and the colored boxes are the new ones.
We implemented maneuver coordination by implementing the MCM on the OpenC2X side and integrating it with Autoware. 
On the OpenC2X side, we implemented the application interface to receive information from Autoware and the MC service to manage states and messages.
On the Autoware side, we extended Autoware's planning function to format the lane-change detection information and trajectory into a form that OpenC2X can use. We also implemented functions for collision detection and calculation, verification, and loading of the prescribed trajectory necessary for generating the prescribed trajectory.

When the \emph{scenario trigger} detects the turn signal from the \emph{behavior planner}, it switches to the advertising phase. 
\emph{Trajectory conversion} converts the trajectory from continuous paired sequences of the location and speed to that of the location and time.
Section~\ref{subsec:conversion} shows the detailed algorithm of the transformation. 
The planned trajectory is transmitted as an advertisement message via OpenC2X. Upon advertisement reception, the receiver returns the intention message using the converted trajectory, as in Section~\ref{subsec:conversion}. These two messages require confirmation via Ack. 

After a predefined period, the prescriber changes its state to prescribing.  
The \emph{collision detector} calculates collision risk using the receivers' and their own planned trajectories. Collision detection is determined by checking whether the distance between vehicles at each time is below a threshold. When there are multiple collisions, the prescriber discovers the target vehicle using the algorithm detailed in Section~\ref{subsec:target_discovery}. 
Then, the \emph{prescribed trajectory generator} calculates the prescribed trajectory for the target vehicle using the algorithm shown in Section~ \ref{subsec:prescribed_generation}. The prescriber then sends the prescribed trajectory to the target vehicle. 

The receiver's  \emph{prescribed trajectory verifier} checks whether the trajectory is acceptable. The receiver returns the decision by returning acceptance or cancellation. 
When it is accepted, the \emph{prescribed trajectory loader} loads the trajectory to the \emph{motion planner}. 

\subsection{Trajectory conversion}\label{subsec:conversion}
The \emph{trajectory converter} converts the trajectory from continuous paired sequences of the location and speed used in Autoware to that of location and time. The converted trajectory makes it easier for the receiver to detect collisions. 
First, we search for the point, ${x_0, v_0, t_0}$, in the trajectory that is closest to the coordinates of the current vehicle.
Next, assuming that the speed from the current position to the next position is constant, the time, $t_1$, at which the vehicle arrives at the next position, is obtained from $\frac{(x_1-x_0)}{v_0}+t_0$.
The process is repeated to obtain the time until the end of the trajectory.
Generalizing this, the $n-$th time of the trajectory starting from the current point, ${x_0, v_0, t_0}$, is expressed by Equation ~\ref{eq:convert}.

\begin{equation}
  t_n = \sum_{k=0}^{n}\frac{(x_n-x_{n-1})}{v_{n-1}}+t_0 \; (n\geq 1). \label{eq:convert}
\end{equation}

Because writing the entire trajectory in the message will not fit into one frame, the trajectory points were thinned out to one-fifth size in order to avoid fragmentation.

\subsection{Target vehicle discovery}\label{subsec:target_discovery}

%
When the \emph{collision detector} detects multiple collisions, the prescriber eliminates the vehicles that are not in the target lane by checking the lane's rectangular zone provided by Lanelet2~\cite{Poggenhans2018-fk} with the lane ID. 
In our implementation, the prescriber instructs the deceleration of the leading vehicle in the lane-change scenario. The vehicles behind it autonomously decelerate with the leading vehicle. 
The leading vehicle can be discovered by repeating the following process. 
Autoware maintains the direction of travel, $\vec{n}$, of the vehicles.
We also calculate the vector, $\overrightarrow{x_ix_{i+1}}$, which indicates the relative position of the two vehicles from the positions of each vehicle ($x_0\sim x_n$). 
Then, if the inner product of two vectors, $\vec{n}\cdot \overrightarrow{x_ix_{i+1}}$, is a positive value, we can determine that $x_{i+1}$ is in front; otherwise, $x_i$ is in front. 

\subsection{Prescribed trajectory generation} \label{subsec:prescribed_generation}
\begin{figure}[htb]
  \centering
  \includegraphics[width=\linewidth]{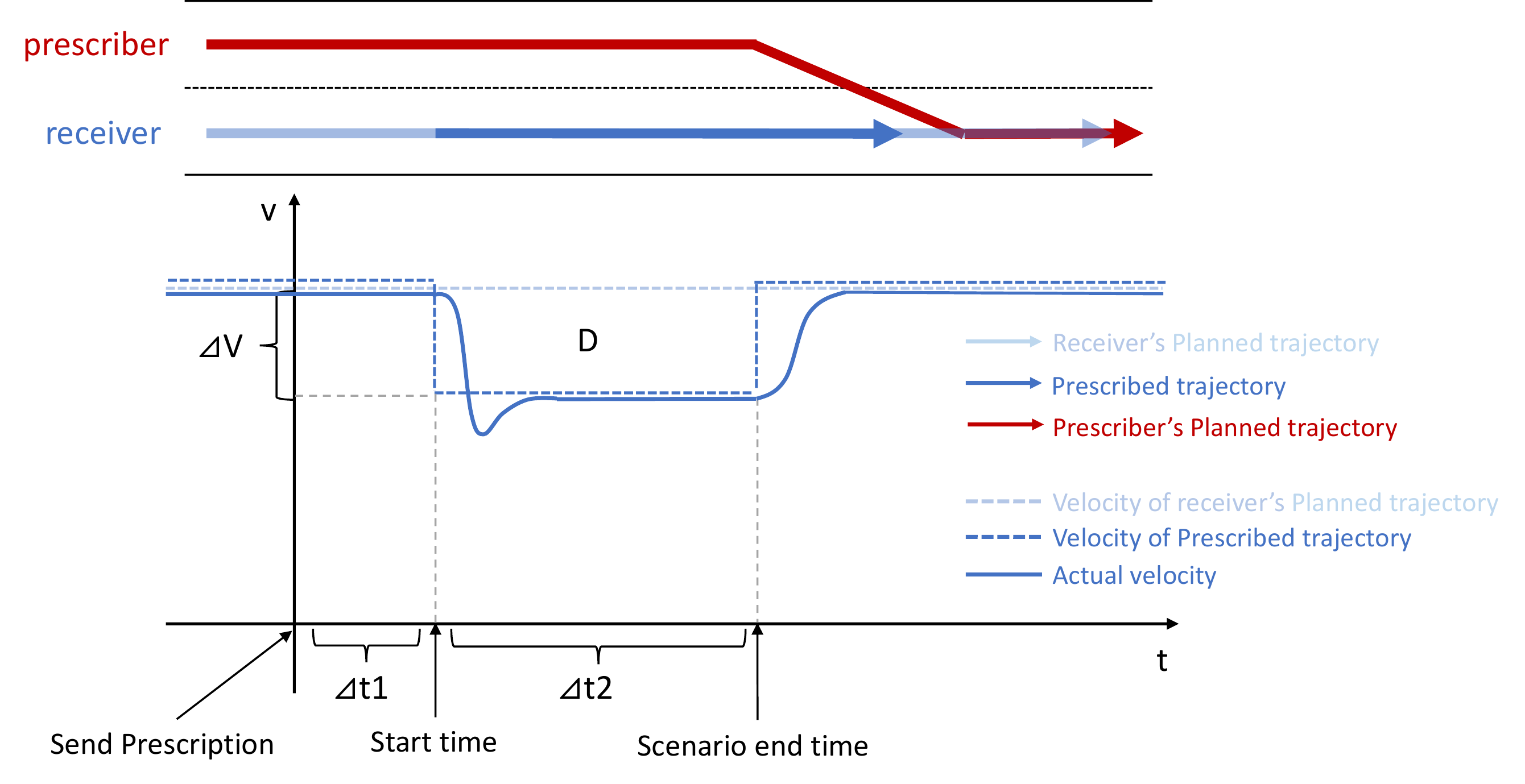}
  \caption{Algorithm for calculating the prescribed trajectory}
  \label{fig:prescribed_trajectory}
\end{figure}

Figure~\ref{fig:prescribed_trajectory} shows the speed of the planned and actual trajectories. In this study, we generated the prescribed trajectory by reducing the receiver's planned trajectory speed. 
First, we set a certain amount of time to wait before starting the action for $\Delta t_1$ from the prescription. During this time, the speed remains constant. 
The system then decelerates at a constant speed, $\Delta V$, during the scenario movement ($\Delta t_2$).
$\Delta V$ is a constant, and $\Delta t_2$ is expressed as $\Delta t_2 = D / V$ using the appropriate vehicle distance, $D$.
Furthermore, $D$ is expressed as $D = d+d_0$ using the current distance, $d$, and the desired distance $d_0$.
After period $\Delta t_2$, it returns to its original speed.
By deceleration, the prescriber obtains a safe distance, $D$, from the receiver.  
The obtained velocity of the prescribed trajectory has a discontinuous value, as indicated by the blue dotted line in Fig.~\ref{fig:prescribed_trajectory}. 
The receiver makes a smooth value of the actual velocity from the prescribed trajectory by feedback control from the current speed, as shown by the solid blue line in the figure.

\section{Evaluation}\label{sec:evaluation}

We performed experiments for maneuver coordination between automated vehicles using MCM in a lane-change scenario on a two-lane road using Autoware’s planning simulator.
We conducted three experiments: measurement of the communication bandwidth consumed in maneuver coordination; arrival time measurement; and the robustness of the system against packet loss. 

\begin{figure}[tbp]
  \centering
  \includegraphics[width=\linewidth]{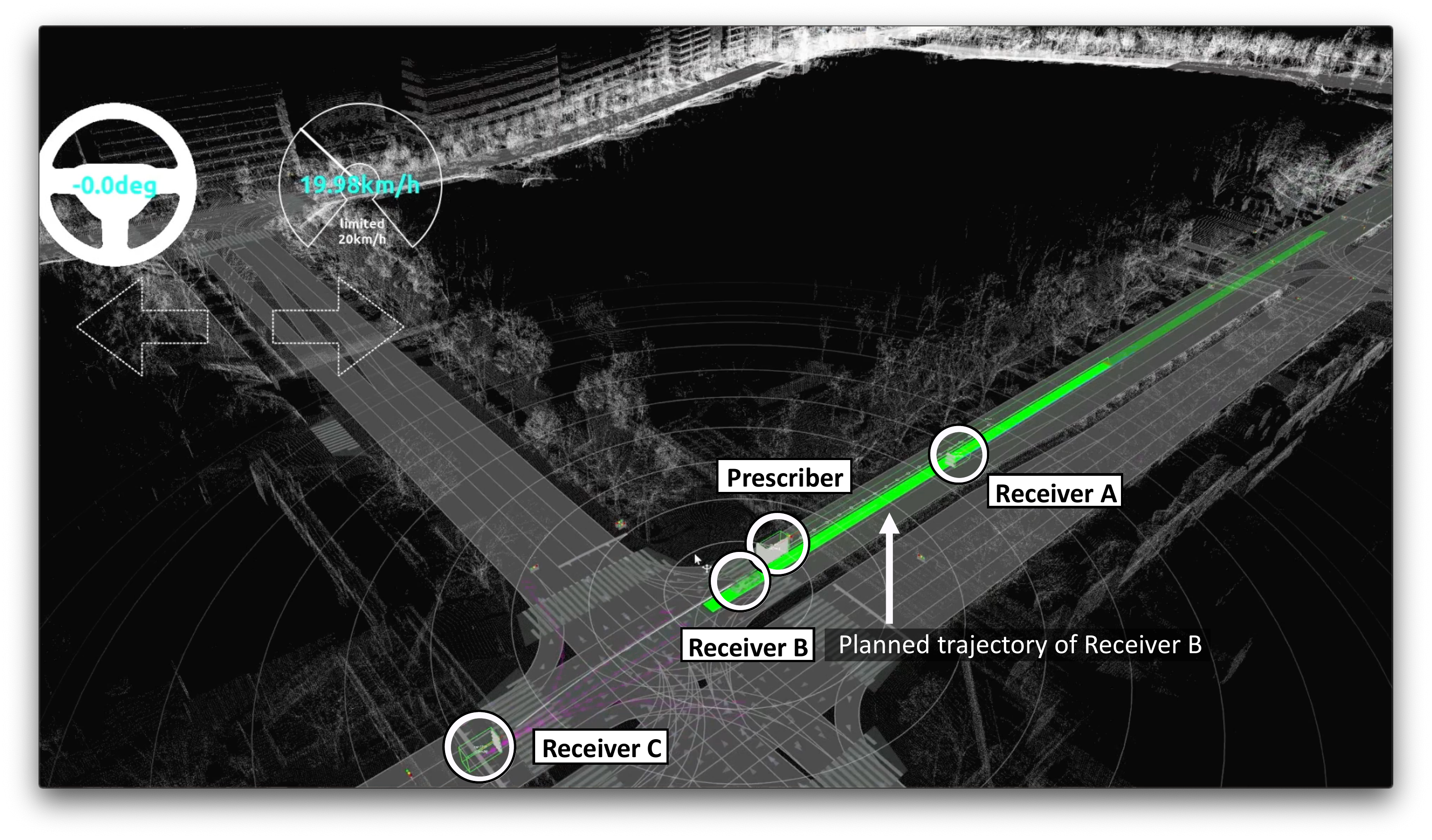}
  \caption{View on RViz of receiver B}
  \label{fig:rviz_view}
\end{figure}

The map used was a straight line in Nishi-Shinjuku, as shown in Fig.~\ref{fig:rviz_view}. 
Figure~\ref{fig:rviz_view} shows the view on RViz of receiver B.
Additionally, the green line represents the planned trajectory of receiver B.
There were start and goal positions. The distance between them was approximately 260 m. 
We installed our implementation (extended Autoware and OpenC2X) on four computers and connected them via Ethernet.

Table \ref{tab:ex_variables} summarizes the values of the important parameters for the experiment. 
First, we provided the timeout duration ($t_{timeout}$) for all messages, set to 2\si{s}. For the robustness experiment, we varied this parameter to 0, 1, and 2\si{s} to measure the impact. The time to resend ($\Delta t_{resend}$) was set to 0.1\si{s}, and the generation frequency ($f$) was set to 10 Hz.
Next, we set the parameters for the presented trajectory.
Prescribed vehicle distance ($D_{prescribed}$) was set to 20 m plus the current vehicle distance.
The deceleration width ($\Delta V$) was 20 km/h, the period from prescription to action, $\Delta t_1$, was the same as $t_{timeout}$, and the period from the action to scenario end time $\Delta t_2$ was obtained by the formula, $D/\Delta V = 3.6$.


\begin{table}[htb]
        \centering
        \caption{\label{tab:ex_variables}Experimental parameters}
        \begin{tabular}{c|cc} \hline
          Type & Valiable name & Value \\ \hline \hline
          \multirow{3}{*}{MCM} & Time to timeout ($t_{timeout}$) & 0\si{s}, 1.0\si{s}, 2.0\si{s}\\
          & Time to resend ($\Delta t_{resend}$) & 0.1s \\
          & Generation frequency ($f$) & 10 Hz \\ \hline
          &
          \begin{tabular}{c}
          Prescribed vehicle distance \\
          ($D_{prescribed}$)
          \end{tabular} & 20m + $d_0$ \\
          Prescribed  & Deceleration width ($\Delta V$) & 20 km/h \\
          trajectory & Prescription to action ($\Delta t_1$) & $t_{timeout}$ \\
          & Action to scenario end ($\Delta t_2$) & $D/\Delta V = 3.6$s \\ \hline
        \end{tabular}
    \end{table}

\subsection{Communication bandwidth measurement}

First, we tested maneuver coordination using MCM with four vehicles. 
Here, a vehicle prescriber changes lanes and three receivers (i.e., A, B, and C) at the lane-change destination. 

\begin{figure}[tbp]
  \centering
  \includegraphics[width=\linewidth]{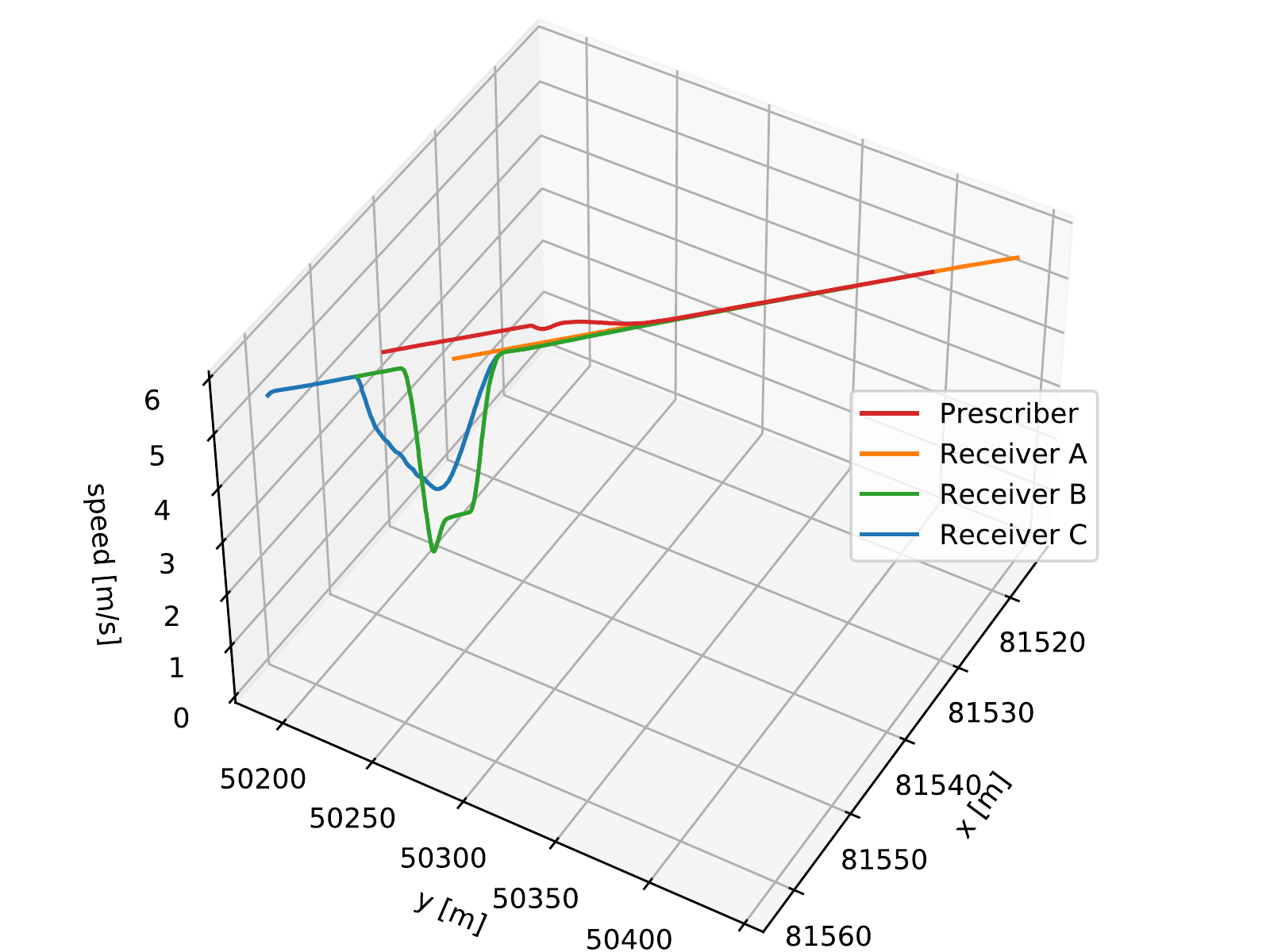}
  \caption{Speed and position of the four vehicles}
  \label{fig:pose_v_multi}
\end{figure}

The prescriber selects an appropriate vehicle from among the receivers and provides instructions to slow down.
Figure~\ref{fig:pose_v_multi} shows the speed of the four vehicles at each position. 
Figure~\ref{fig:pose_v_multi} shows that receiver B decelerates prior to the prescriber's lane change. 
This shows that the following vehicle (receiver C) also decelerates to maintain an appropriate distance from receiver B. 

\begin{figure}[htb]
  \begin{minipage}[b]{\linewidth}
    \centering
    \includegraphics[width=\linewidth]{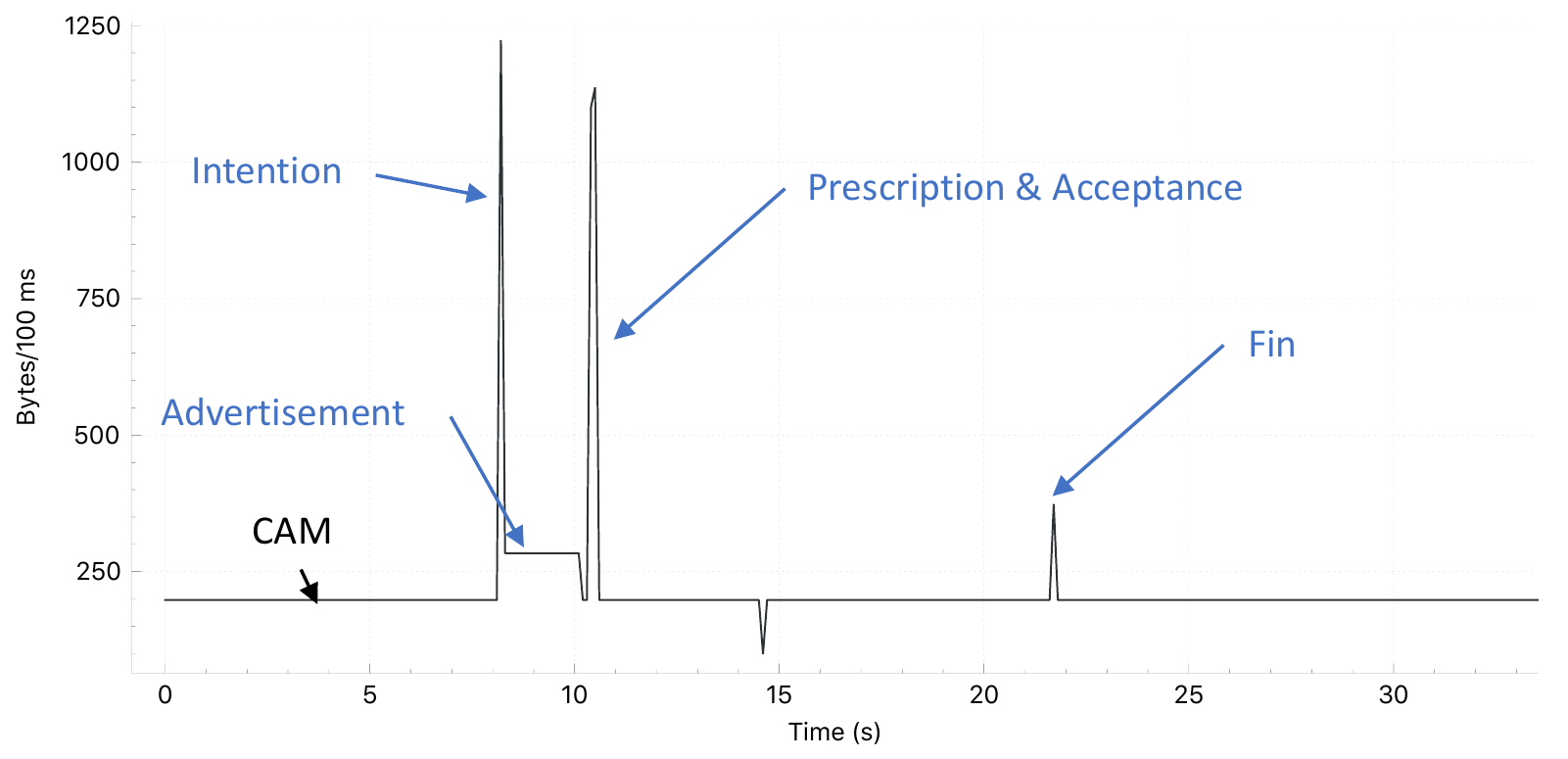}
    \subcaption{Two vehicles}
    \label{fig:bandwidth_2}
  \end{minipage} \\
  \begin{minipage}[b]{\linewidth}
    \centering
    \includegraphics[width=\linewidth]{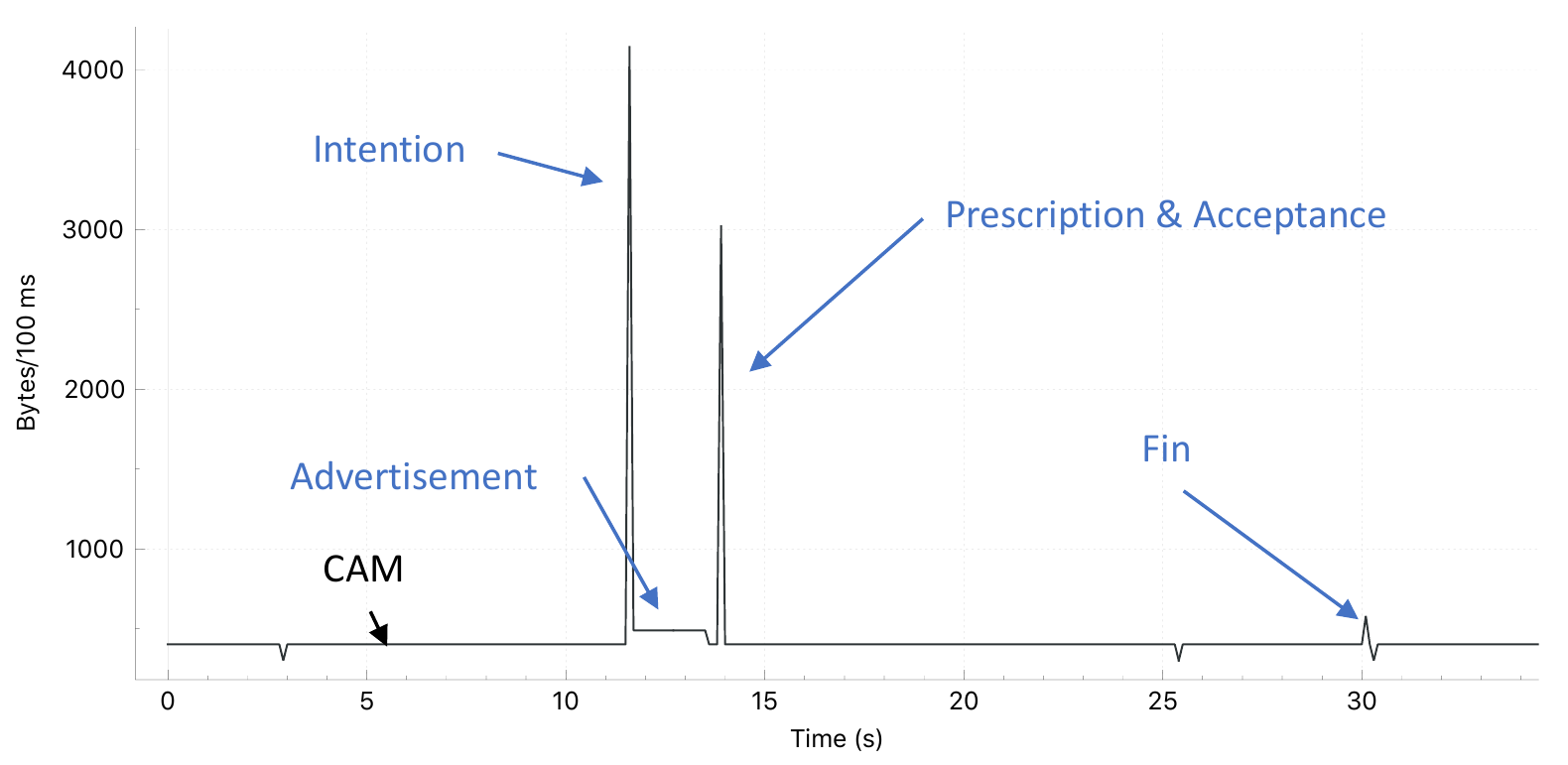}
    \subcaption{Four vehicles}
    \label{fig:bandwidth_4}
  \end{minipage}
  \caption{Communication volume at each time}
  \label{fig:bandwidth}
\end{figure}

Figure~\ref{fig:bandwidth_2} shows the communication volume with two vehicles, and Fig.~\ref{fig:bandwidth_4} shows that of four vehicles.
In this experiment, both MCM and CAM were transmitted at 10 Hz for comparison.
Figure~\ref{fig:bandwidth} shows that the MCM containing the trajectory was more than 10-times larger than the CAM.
Thus, the reduction of MCMs is important for bandwidth savings.
Comparing Fig.~\ref{fig:bandwidth_2} and Fig.~\ref{fig:bandwidth_4}, we can see that the intention in Fig.~\ref{fig:bandwidth_2} was three times higher, whereas the amounts of prescription, acceptance, and fin were the same.
This is because receivers A, B, and C sent Intentions to the prescriber, while the prescriber only sent a prescription to receiver B; it also sent a cancel to receivers A and C.
From the results, we observe that the constant delivery of intentions by all vehicles led to bandwidth pressure, and our proposed method of event-driven messaging effectively saved bandwidth.

\subsection{Arrival-time measurement}
\begin{figure}[htb]
  \begin{minipage}[b]{\linewidth}
    \centering
    \includegraphics[width=0.8\linewidth]{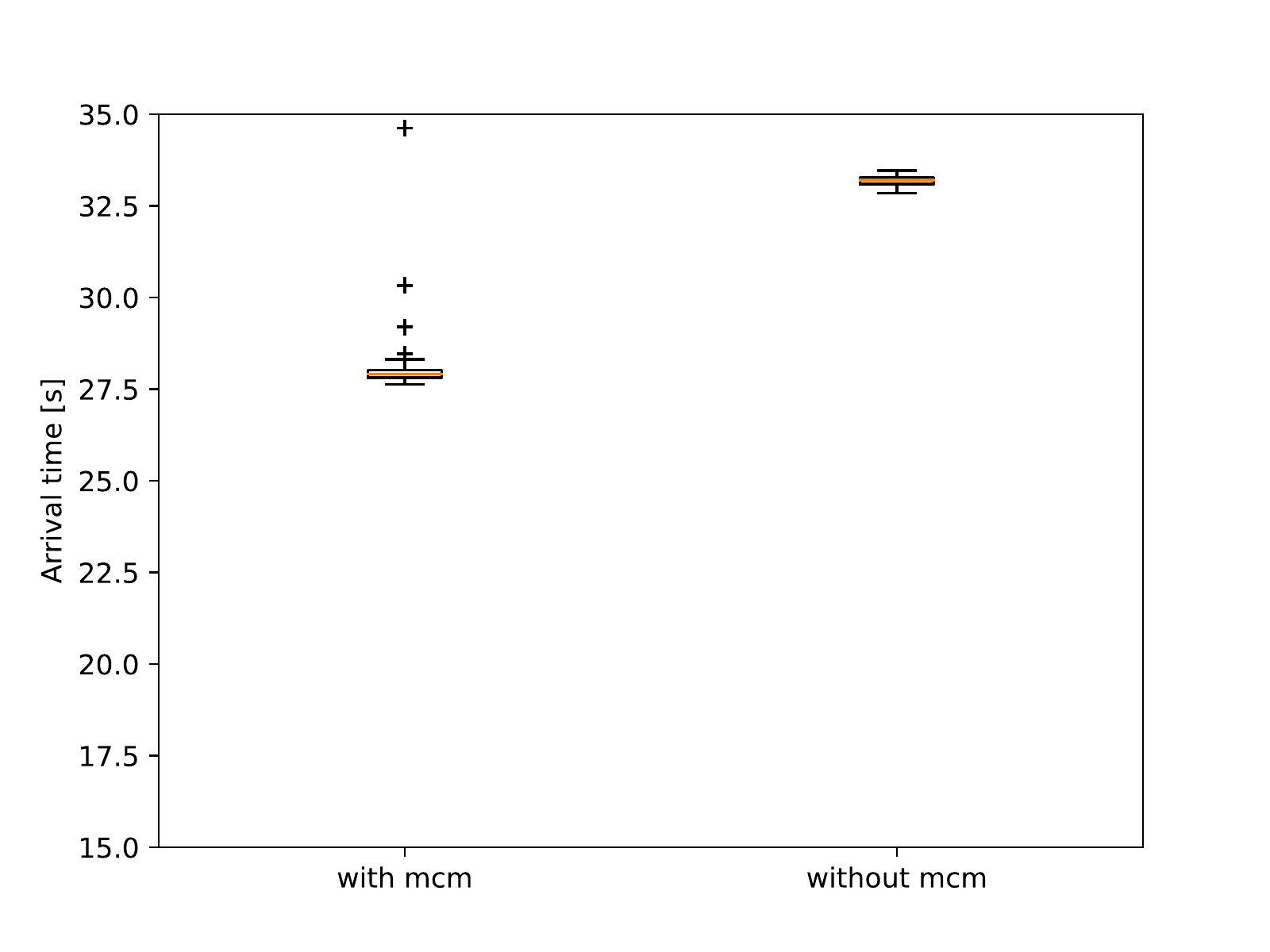}
    \subcaption{Vehicles' speed was 30 km/h.}
    \label{fig:goaltime_30}
  \end{minipage} 
  \begin{minipage}[b]{\linewidth}
    \centering
    \includegraphics[width=0.8\linewidth]{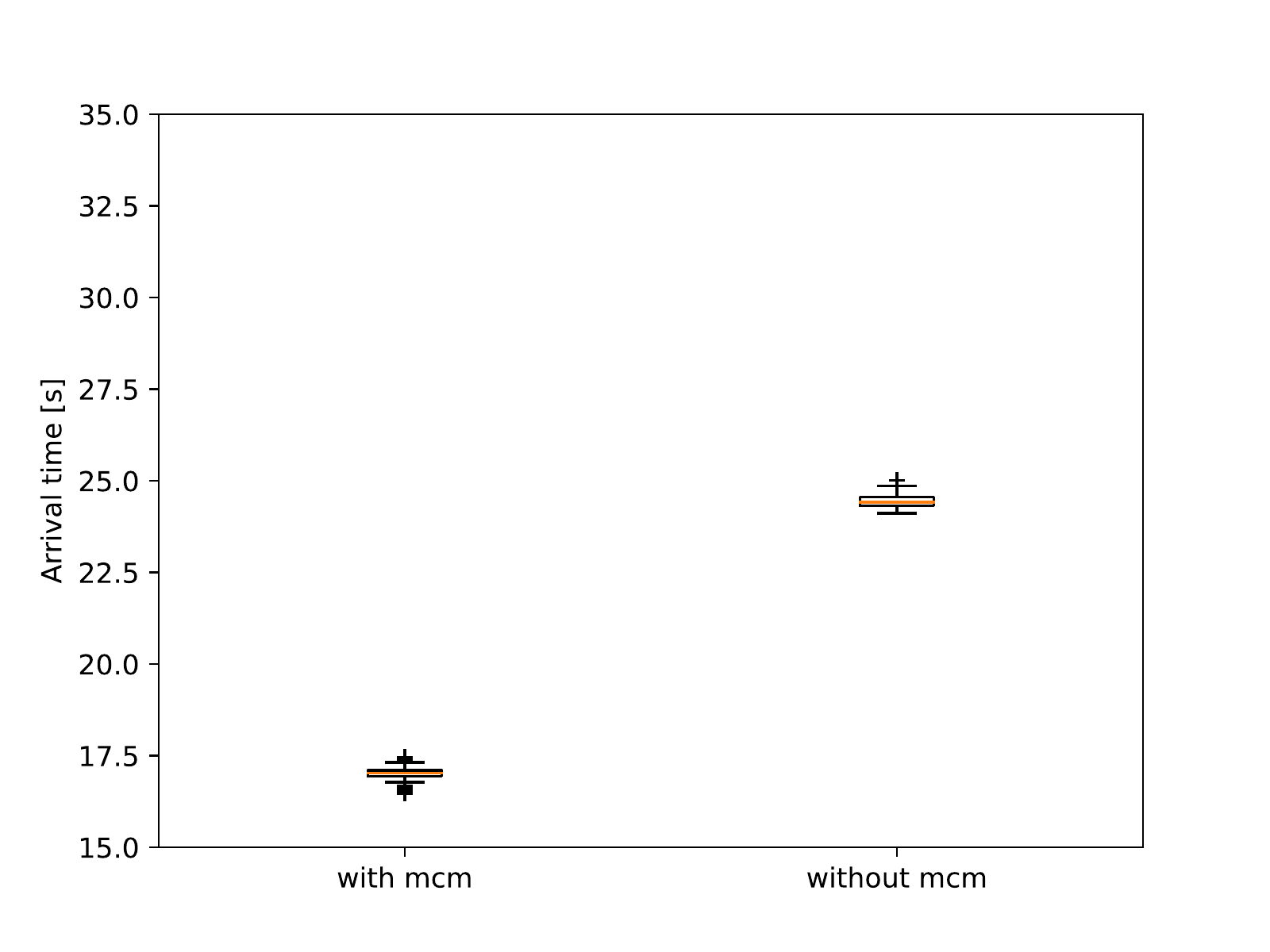}
    \subcaption{Vehicle speed was 50 km/h.}
    \label{fig:goaltime_50}
  \end{minipage}
  \caption{Time from when the prescriber detects a lane change to when the receiver arrives at the target point.}
  \label{fig:goaltime}
\end{figure}

We evaluated the arrival time to 260-m away in the scenario in which one vehicle (prescriber) changed lanes, and another vehicle (receiver) was in the destination lane.
We measured the arrival times for the speeds of two vehicles at 30 and 50 km/h.
Figure \ref{fig:goaltime} shows the results of 100 trials. 
The results show that the arrival times were approximately 5\si{s} faster (15\% faster) when the vehicle speed was 30 km/h and approximately 7\si{s} faster (28\% faster) when the vehicle speed was 50 km/h using MCM. 

The behavior difference with and without MCM affected the result. 
When the MCM was not used, the vehicle did not notice the lane change until just before it happened, detecting a collision only after the lane change started. Then, the vehicle decelerated rapidly.
The vehicle stopped after the sudden deceleration and started again when it secured a sufficient distance between vehicles.
On the other hand, when the MCM was used, the vehicle decelerated to obtain a sufficient distance before changing lanes.
Therefore, when changing lanes, the vehicle traveled with little or no deceleration.

\subsection{Robustness against packet loss}
The probability of successful maneuver coordination increases with the time to timeout ($t_{timeout}$) because the number of message retransmissions increases. 
On the other hand, if $t_{timeout}$ is too long, the time required for maneuver coordination increases, owing to many message retransmissions under a high packet-loss ratio. 
Additionally, the probability of the surrounding situation change increases during the repeated retransmissions.
Therefore, the timeout period should be as short as possible within the range in which the maneuver coordination can operate smoothly. We sought the optimal $t_{timeout}$ experimentally.

\begin{figure*}[htbp]
  \begin{tabular}{ccc}
    \begin{minipage}[b]{0.33\linewidth}
      \centering
      \includegraphics[width=\linewidth]{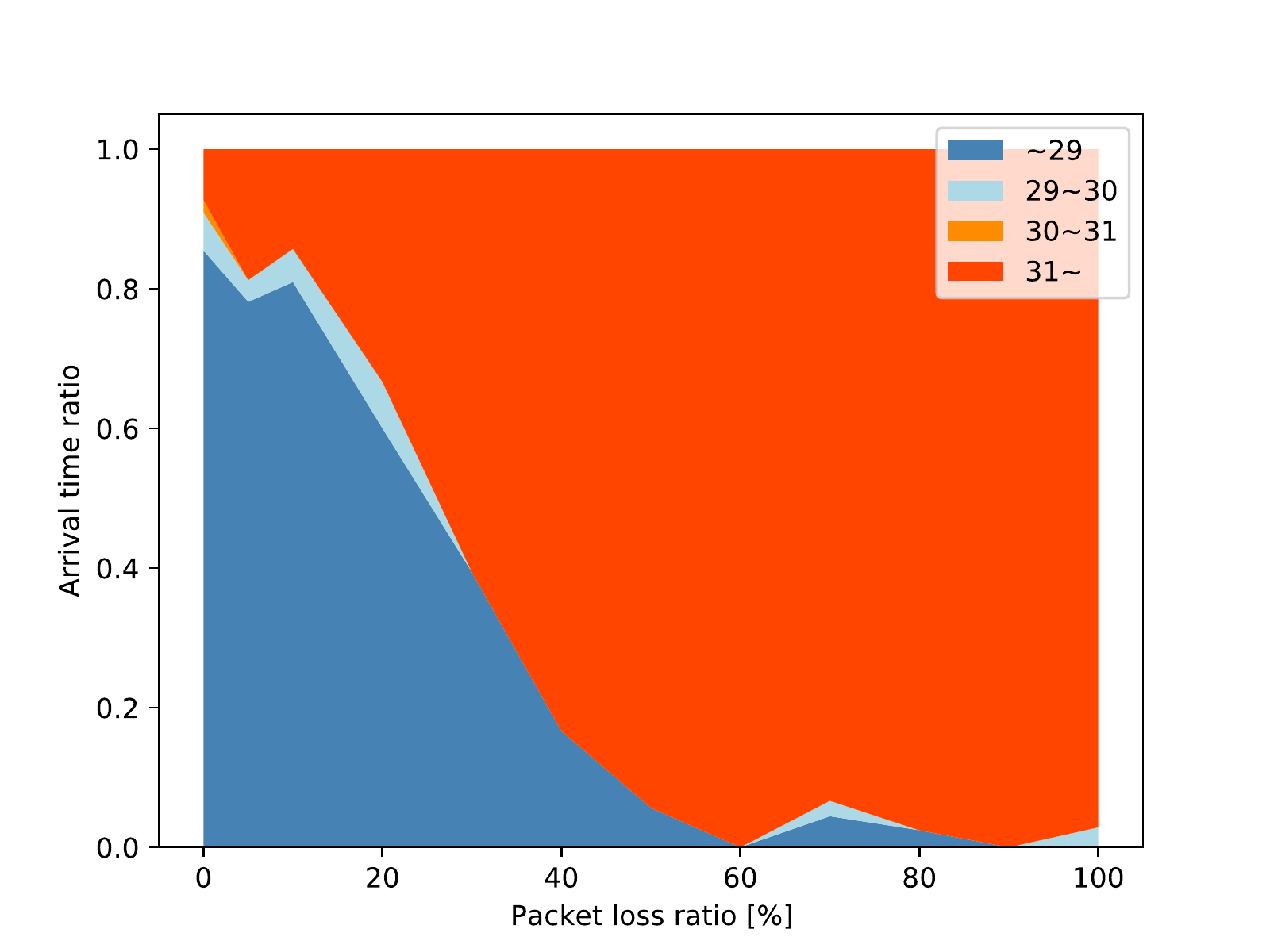}
      \subcaption{Time to timeout was $0\si{s}$.}
      \label{fig:stack_loss_0}
    \end{minipage}
    \begin{minipage}[b]{0.33\linewidth}
      \centering
      \includegraphics[width=\linewidth]{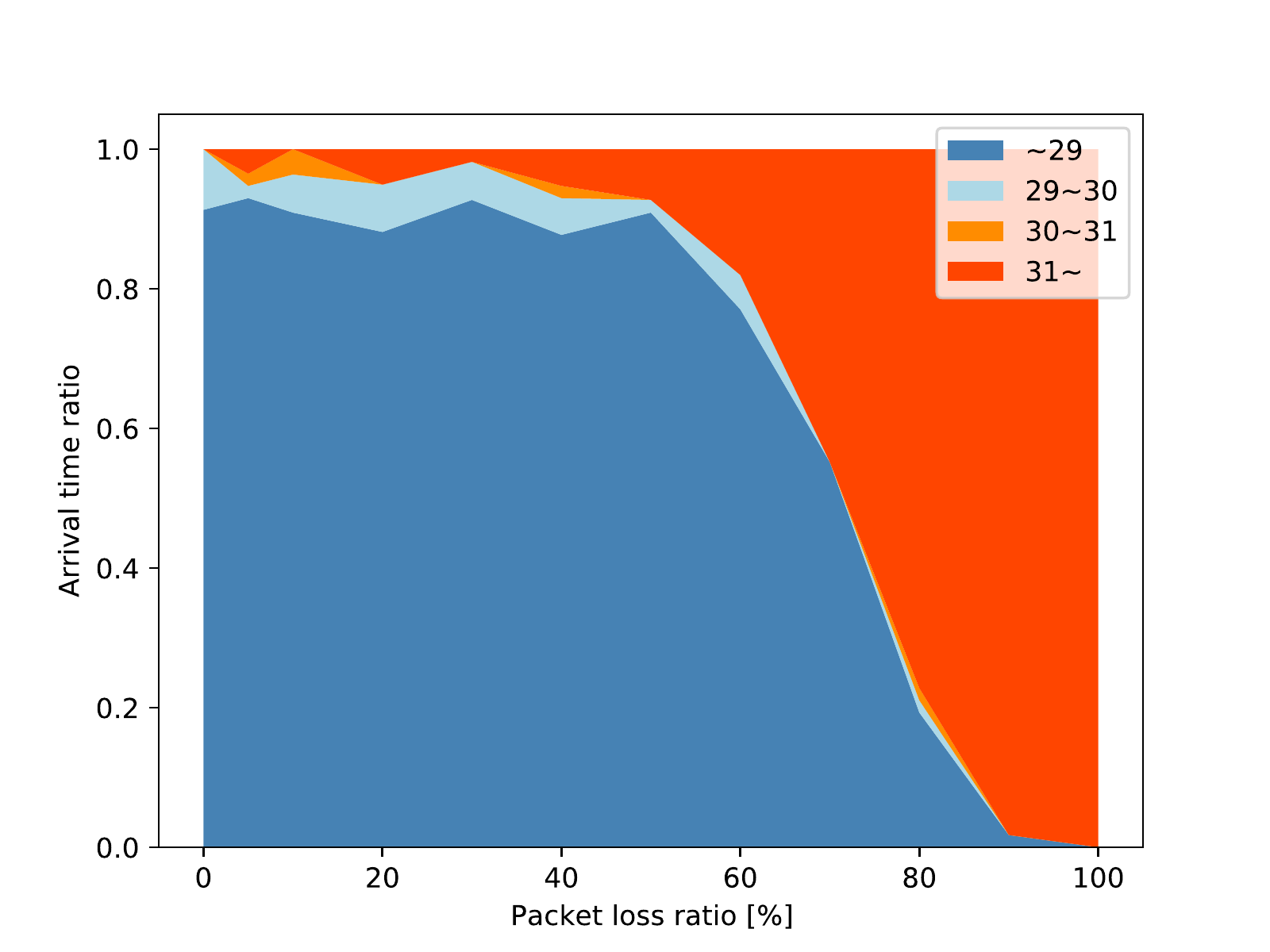}
      \subcaption{Time to timeout was $1\si{s}$.}
      \label{fig:stack_loss_1}
    \end{minipage}
    \begin{minipage}[b]{0.33\linewidth}
      \centering
      \includegraphics[width=\linewidth]{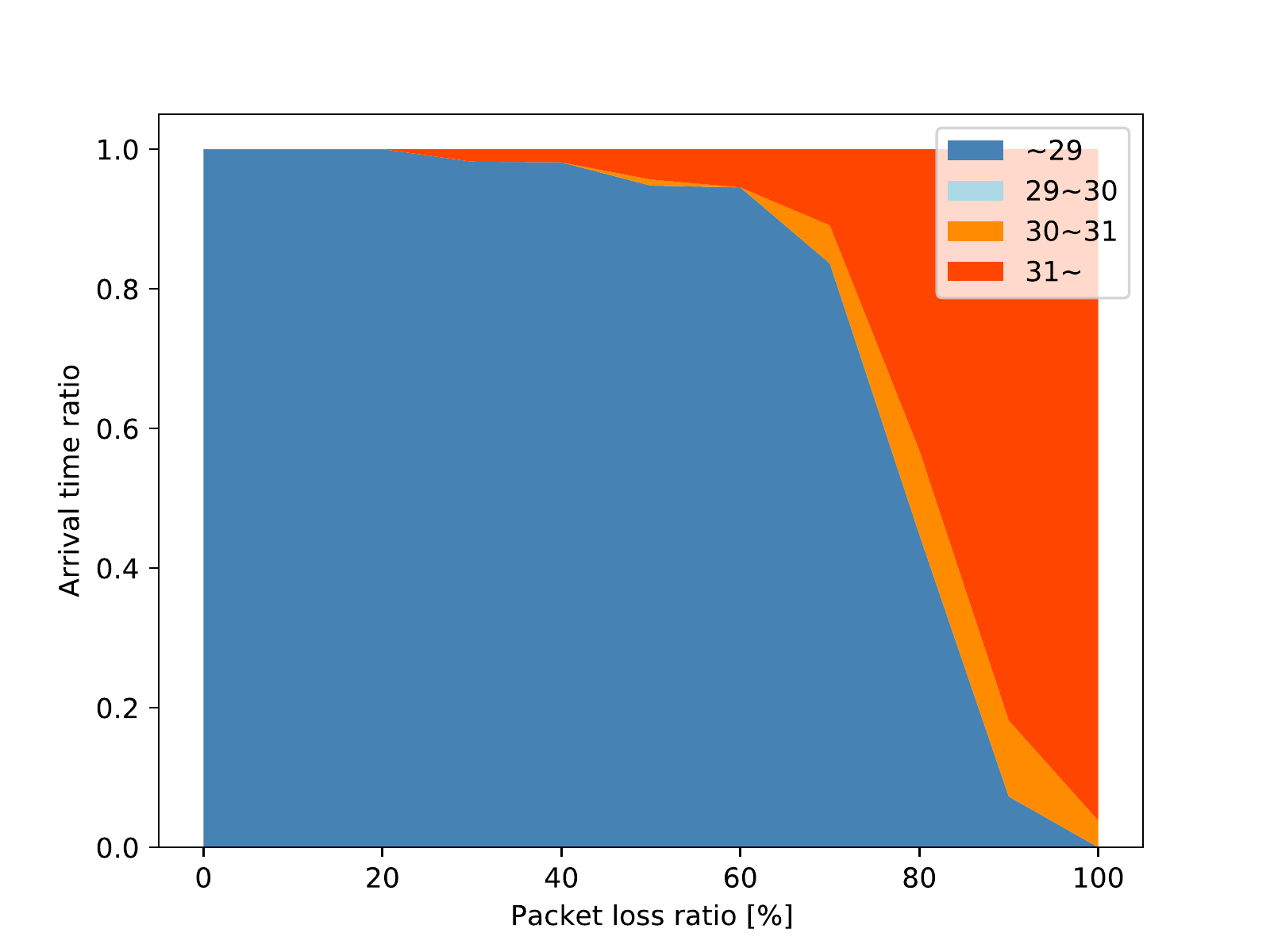}
      \subcaption{Time to timeout was $2\si{s}$.}
      \label{fig:stack_loss_2}
    \end{minipage}
  \end{tabular}
  \caption{Percentage of time from when the prescriber detects a lane change to when the receiver arrives at the target point.}
  \label{fig:stack_arrival_time}
\end{figure*}

We performed the arrival-time measurement by varying the packet-loss rate from 0 to 100\% at a 10\% interval. We conducted the measurement by changing the timeout period of each message, $t_{timeout}$, to $0, 1$, and $2\si{s}$. The experiments were conducted 60 times. In the experiment, two vehicles ran at 30 km/h and performed lane-change coordination. 
Figure \ref{fig:stack_arrival_time} shows the arrival time in percentages, classified into 29\si{s} or less, 29--30\si{s}, 30--31\si{s}, and 31\si{s} or more.

From the figure, we can see that the arrival time increased rapidly from 10\% at $t_{timeout}= 0\si{s}$, from 60\% at $t_{timeout}= 1\si{s}$, and from 70\% at $t_{timeout}= 2\si{s}$. From the result, $t_{timeout}=2$ had the most robust maneuver coordination against packet loss. 
The probability of successful transmission of a message, $p$, was given by Equation ~\ref{eq:probability_of_success}: 
\begin{equation}
  p=1-\lambda^{t_{timeout}/t_{resend}},  \label{eq:probability_of_success}
\end{equation}
where $\lambda$ denotes the packet loss rate, $t_{timeout}$ is the time to timeout, and $t_{resend}(=0.1\si{s})$ denotes the time until retransmission. 
For example, when $t_{timeout}$ is 1\si{s}, the value of $\lambda^{t_{timeout}/t_{resend}}$ is $\lambda^{10}$.
Therefore, if $\lambda$ is a small value, $\lambda^{10}$ will be negligible, but $\lambda^{10}$ will increase exponentially according to $\lambda$.
Because the probability of maneuver coordination is proportional to the probability of $p$, the arrival time sharply increases as the packet-loss ratio increases. 
Additionally, according to a previous practical study of vehicle-to-vehicle communication, \cite{Tsukada2014, Tsukada2014b}, the packet loss ratio can be kept below 20\% in the absence of obstacles, such as buildings and trees. Therefore, we conclude that our system is robust against packet loss when the timeout is $1$ or $2\si{s}$. 

\section{Conclusions}\label{sec:conclusions}

In this paper, we discussed the issues and requirements for maneuver coordination in autonomous driving. We proposed an MCM protocol that satisfies these requirements using seven types of messages with state management. 
We also divided it into two parts, service and application, which are common and unique to various scenarios and applications to increase the versatility of the protocol. 
We implemented AutoMCM for realizing the proposed MC protocol by extending Autoware and OpenC2X.
In an experiment with four vehicles, the proposed event-driven message exchange effectively reduced the communication bandwidth by limiting the MCM transmission, which was 10-times larger than the message size of CAM. In the arrival-time measurement, we observed that our proposed method achieved 15\% faster performance when the vehicle speed was 30 km/h and 28\% faster when the vehicle speed was 50 km/h. Additionally, our system showed robustness against packet loss when the timeout was set to $1$ or $2\si{s}$.

Future work includes implementing other scenario applications, such as intersections, ramp merging, and pedestrian crossing. Furthermore, it will be necessary to evaluate whether our scheme can coordinate maneuvers in more than two lanes. 
We also plan to evaluate our implementation with field experiments. 
In this study, we used a wired network. However, we plan to use a wireless network or network simulator to reproduce more realistic network characteristics. 

\section*{Acknowledgment}

This work was partly supported by JSPS KAKENHI (grant number: 19KK0281 and 21H03423). 

\bibliographystyle{unsrt}
\bibliography{main.bib}

\end{document}